\pdfoutput=1

\documentclass[11pt]{article}
\usepackage[table,usenames,dvipsnames]{xcolor}
\usepackage{enumitem}

\usepackage{algorithm}
\usepackage{algpseudocode}
\usepackage{coling}
\usepackage{tabularx}
\usepackage{comment}
\usepackage{placeins}
\usepackage{stfloats} 
\usepackage{placeins}
\usepackage{float} 
\usepackage{tabu} 
\usepackage{multirow}
\usepackage{booktabs}
\usepackage{hyperref}
\usepackage{times}
\usepackage{latexsym}
\usepackage{graphicx}
\usepackage{amsmath}
\usepackage{threeparttable}
\usepackage{enumitem}
\usepackage{booktabs}
\usepackage{amssymb}
\usepackage{subcaption}
\usepackage{float}
\usepackage{url}
\usepackage{hyperref}

\usepackage{etoolbox} 

\renewenvironment{quote}
  {\list{}{\rightmargin=0.2cm \leftmargin=0.2cm}%
   \item\relax}
  {\endlist}

\usepackage{tikz}
\usepackage{array}
\usepackage{tabularx}
\usepackage{cellspace}
\usetikzlibrary{shapes,arrows,positioning}
\definecolor{ao(english)}{rgb}{0.0, 0.5, 0.0}
\definecolor{brass}{rgb}{0.6, 0.8, 0.2}
\definecolor{chromeyellow}{rgb}{1.0, 0.65, 0.0}
\definecolor{crimson}{rgb}{0.86, 0.08, 0.26}

\usepackage{titlesec}

\titlespacing*{\paragraph}{%
  0pt}{
  0.3\baselineskip}{
  1em}

\usepackage{times}
\usepackage{latexsym}
\usepackage{float} 
\usepackage[T1]{fontenc}

\usepackage[utf8]{inputenc}

\usepackage{microtype}

\usepackage{inconsolata}
\usepackage{pgfplots}
\pgfplotsset{width=10cm,compat=1.17}
\usepackage{pgf-pie}
\usepackage[most]{tcolorbox}
\usepackage{enumitem}
\usepackage[absolute,showboxes]{textpos} 
\textblockorigin{0pt}{0pt} 
\TPshowboxestrue 
\TPshowboxesfalse 
\usetikzlibrary{pgfplots.polar, positioning}

\newtcolorbox{mybox}[2][]{enhanced, 
    colback=blue!2!white, colframe=blue!75!black, 
    fonttitle=\bfseries, coltitle=white, title=#2,
    sharp corners=south, rounded corners=north,
    boxrule=0.4mm, boxsep=5pt,
    borderline={0.3mm}{0.3mm}{gray!50!black},
    drop shadow={white, opacity=0.5, xshift=2.5mm, yshift=-2.5mm},
    attach boxed title to top left={yshift=-3mm, xshift=5mm},
    boxed title style={colback=blue!75!black, sharp corners},
    #1}

\definecolor{pastelgreen}{rgb}{0.75, 1.0, 0.85}
\definecolor{pastelblue}{rgb}{0.85, 0.85, 1.0}
\definecolor{pastelpink}{rgb}{1.0, 0.85, 0.9}
\definecolor{pastelyellow}{rgb}{1.0, 1.0, 0.8}
\definecolor{pastelpurple}{HTML}{D1CCEA} 
\definecolor{pastelorange}{rgb}{1.0, 0.9, 0.75}
\definecolor{pastelred}{rgb}{1.0, 0.8, 0.8}
\definecolor{pastelgray}{rgb}{0.9, 0.9, 0.9}
\definecolor{pastelcyan}{rgb}{0.8, 0.95, 1.0}
\definecolor{pastelbrown}{rgb}{0.95, 0.85, 0.75}
\definecolor{pastelpurplebox}{rgb}{0.75, 0.58, 0.89}

\newcolumntype{L}[1]{>{\raggedright\let\newline\\\arraybackslash\hspace{0pt}}m{#1}}
\newcolumntype{C}[1]{>{\centering\let\newline\\\arraybackslash\hspace{0pt}}m{#1}}
\newcolumntype{R}[1]{>{\raggedleft\let\newline\\\arraybackslash\hspace{0pt}}m{#1}}

\newcommand{\corpusname}[0]{\textsc{NYT-Connections}}

\title{%
    \begin{textblock*}{1cm}[0,0](2.7cm,2.1cm)  
    \end{textblock*}
    \corpusname{:} A Deceptively Simple Text Classification Task \\ that Stumps System-1 Thinkers
}

  \author{
    Angel Yahir Loredo Lopez$^1$\textnormal{,} 
    Tyler McDonald$^2$\textnormal{, and} 
    Ali Emami$^2$\\
    $^1$Universidad Autónoma de San Luis Potosí, San Luis Potosí, Mexico\\
    $^2$Brock University, Saint Catharines, Canada \\
    \texttt{a327322@alumnos.uaslp.mx}, \texttt{\{tmcdonald3, aemami\}@brocku.ca} \\
}

\begin{document}

\maketitle

\begin{abstract}
Large Language Models (LLMs) have shown impressive performance on various benchmarks, yet their ability to engage in deliberate reasoning remains questionable. We present \corpusname{}, a collection of 358 simple word classification puzzles derived from the New York Times Connections game. This benchmark is designed to penalize quick, intuitive ``System 1'' thinking, isolating fundamental reasoning skills. We evaluated six recent LLMs, a simple machine learning heuristic, and humans across three configurations: single-attempt, multiple attempts without hints, and multiple attempts with contextual hints. Our findings reveal a significant performance gap: even top-performing LLMs like GPT-4 fall short of human performance by nearly 30\%. Notably, advanced prompting techniques such as Chain-of-Thought and Self-Consistency show diminishing returns as task difficulty increases. \corpusname{} uniquely combines linguistic isolation, resistance to intuitive shortcuts, and regular updates to mitigate data leakage, offering a novel tool for assessing LLM reasoning capabilities.\footnote{The \corpusname{} dataset is publicly available \href{https://huggingface.co/datasets/tm21cy/NYT-Connections}{here}, with updates to include 28-31 new puzzles monthly.}

\end{abstract}

\section{Introduction}

As Large Language Models (LLMs) continue to advance, the need for effective benchmarks to assess their true capabilities has become increasingly important. While numerous natural language tasks and datasets have been developed across domains such as text summarization, commonsense reasoning, and question answering \citep{hendrycks2021measuringmassivemultitasklanguage, cobbe2021trainingverifierssolvemath, hendrycks2021measuringmathematicalproblemsolving}, these benchmarks often fall short in isolating and evaluating specific cognitive abilities.

One major challenge is the difficulty in assessing individual model capabilities independently. Many existing tasks combine multiple cognitive processes, making it challenging to evaluate distinct abilities \citep{gema2024we, gautam2024robust}. For instance, tasks that simultaneously require mathematical reasoning, natural language understanding, and contextual disambiguation \citep{patel-etal-2021-nlp} can obscure a model's true proficiency in any single area.

Furthermore, many current benchmarks are vulnerable to shortcuts or heuristics. Models may exploit statistical regularities or superficial cues rather than demonstrating genuine understanding, a phenomenon known as `shortcut learning' \citep{geirhos2020shortcut,trichelair-etal-2019-reasonable}. This issue is closely related to the distinction between System 1 and System 2 thinking, as described by \citet{Hagendorff_2023}:

\vspace{-1mm}
\begin{quote}
``\textbf{System 1} processes are fast, automatic and instinctual. They often involve heuristics, or mental shortcuts, which enable quick judgments and decisions without conscious effort. [...] \textbf{System 2} processes, on the other hand, are deliberate and require conscious effort.''
\end{quote}
\vspace{-1mm}

Consequently, many current benchmarks inadvertently reward System 1-style thinking, allowing models to achieve high scores without demonstrating the deliberate reasoning we aim to evaluate.

Finally, as LLMs are trained on increasingly vast amounts of data, the risk of test set leakage into training data grows, potentially leading to inflated performance metrics that do not reflect true generalization capabilities \citep{balloccu-etal-2024-leak, huang2024competition}.

To address these challenges, we introduce \corpusname{}, a novel benchmark of 358 puzzles derived from the New York Times' Connections game. This task requires grouping 16 interrelated terms into 4 sets of 4 closely related words, deliberately tempting incorrect System 1 responses while requiring System 2 thinking for correct solutions. \corpusname{} offers several key advantages:

\begin{figure*}[ht]
	\centering
	\begin{subfigure}[t]{0.45\textwidth}
		\centering
		\includegraphics[width=\textwidth]{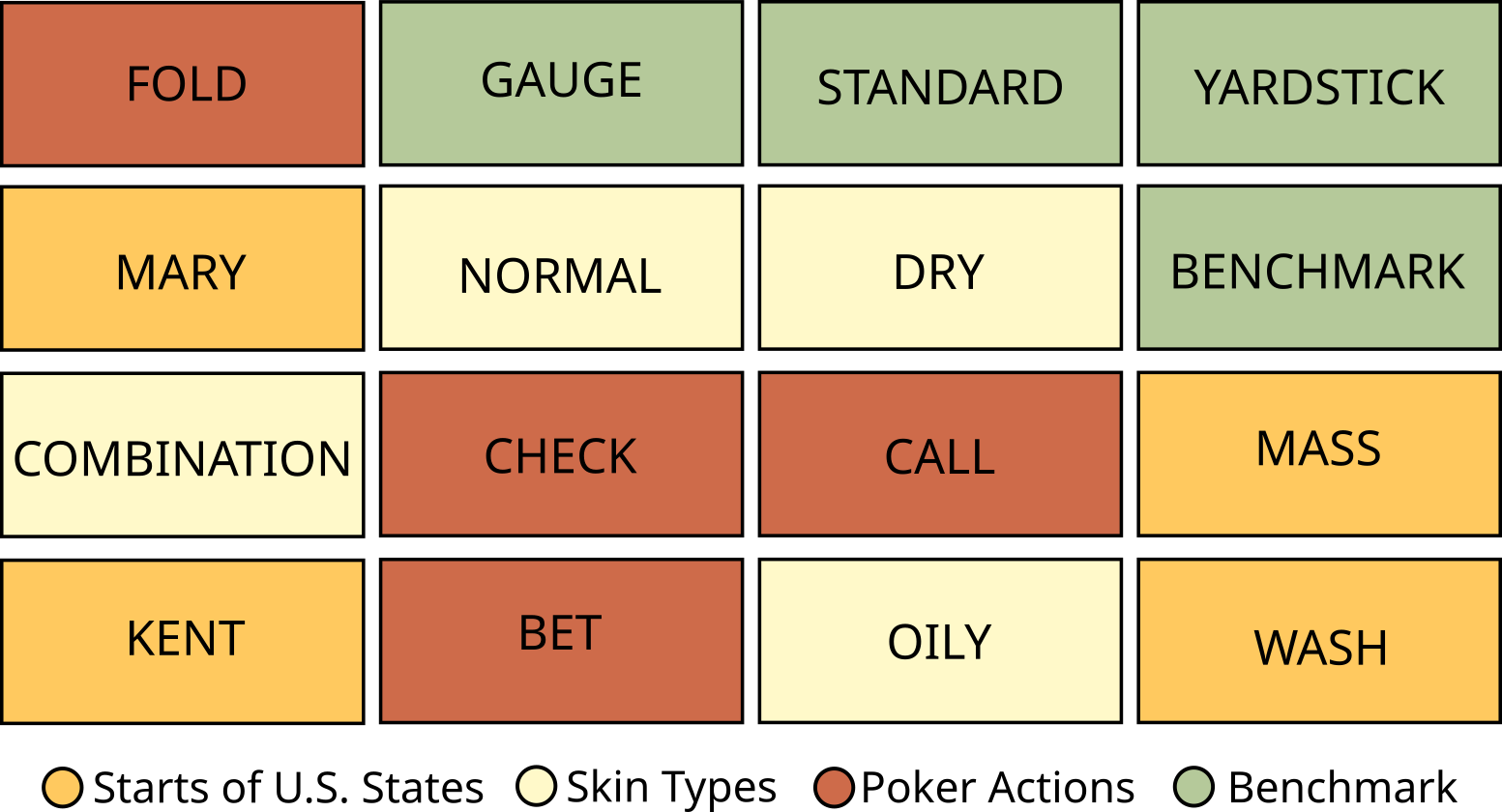}
		\caption{Sample instance of \textit{Connections} showing a grid with various terms; each is marked to indicate its category.}
		\label{fig:game}
	\end{subfigure}
	\hfill
	\begin{subfigure}[t]{0.45\textwidth}
		\centering
		\includegraphics[width=\textwidth]{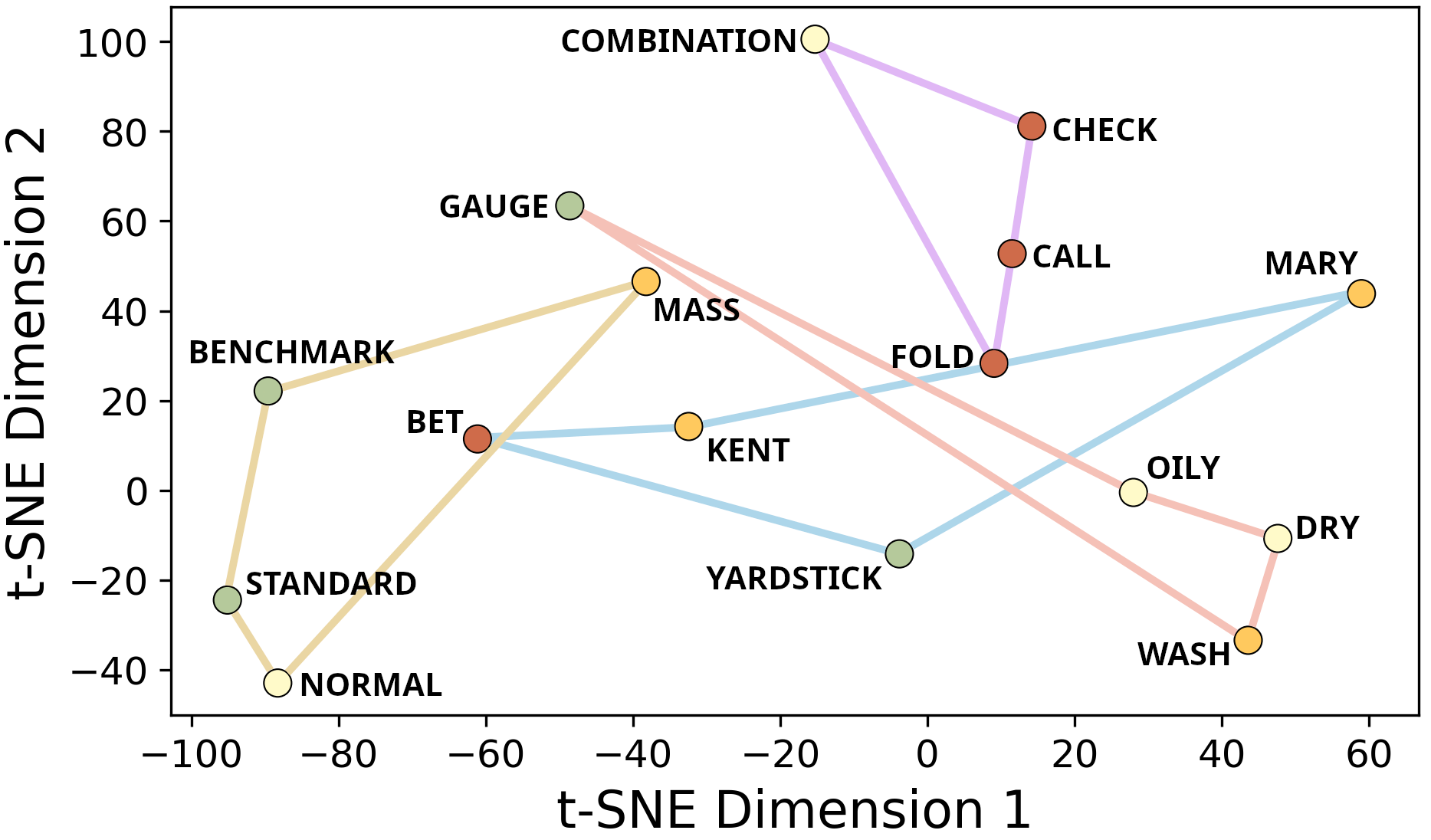}
		\caption{2D t-SNE visualization of term embeddings color-coded by category, illustrating clustering patterns.}
		\label{fig:clusters}
	\end{subfigure}
 \vspace{-1mm}
	\caption{Overview of \textit{Connections} game instance and its embeddings visualization.}
	\label{fig:gameandclusters}
\end{figure*}

\begin{itemize}[itemsep=-1pt, topsep=0pt, parsep=0pt, leftmargin=*]
\item \textbf{Linguistic Isolation:} It focuses purely on word relationships, minimizing confounding factors.
\item \textbf{System 2 Emphasis:} It penalizes quick, intuitive responses and requires deliberate reasoning.
\item \textbf{Continual Novelty:} With daily updates, it provides a stream of novel instances, mitigating data leakage concerns.
\end{itemize}

In this paper, we contribute the following:

\begin{enumerate}[itemsep=-1pt, topsep=0pt, parsep=0pt, leftmargin=*]

\item We present NYT-Connections, a benchmark designed to isolate and evaluate deliberate reasoning in LLMs.

\item We provide a comprehensive evaluation of six recent LLMs, a simple machine learning heuristic, and human performance on NYT-Connections.

\item We analyze various prompting techniques and their effectiveness in promoting System 2 ``reasoning'' in LLMs.

\item We demonstrate a substantial performance gap between LLMs and humans, with even the most advanced models falling short by nearly 30\%.

\end{enumerate}


\section{\corpusname}
\subsection{The Task}

\corpusname{} is based on \textit{Connections}, a word classification game by the New York Times \citep{NYTimes2024}. This daily puzzle challenges players to group 16 terms into 4 sets of 4 related words. Its design intentionally tempts quick, obvious associations but requires careful, deliberate reasoning to solve correctly \citep{NYT2023connections}.

Figure~\ref{fig:game} illustrates this design. The correct ``Skin Types'' group includes ``\textbf{Normal}'', ``Dry'', ``Combination'', and ``Oily". However, ``\textbf{Normal}'', ``Standard'', and ``Benchmark'' temptingly form a ``Status Quo'' group of three. This misdirection is crucial: the apparent ``Status Quo'' grouping leaves only three ``Skin Types'' terms, making it impossible to form four complete groups.

To demonstrate the challenge for machine learning approaches, we applied $k$-means clustering to Multilingual-E5-Large-Instruct word embeddings \citep{wang2024multilinguale5textembeddings}. As shown in Figure~\ref{fig:clusters}, this method fails to correctly classify the terms, instead grouping semantically related words from different categories. 

\subsection{Dataset Construction}
We constructed the \corpusname{} dataset through the following process:

\textbf{Dataset Collection:} We compiled the complete set of 358 Connections puzzles from an archive covering daily offerings from the game's debut on June 12, 2023, to June 3rd, 2024\footnote{\href{https://tryhardguides.com/nyt-connections-answers/}{Connections Puzzles Archive}}.

\textbf{Difficulty Assessment:} To gauge the perceived difficulty of each puzzle, we cross-referenced our dataset with an existing difficulty chart\footnote{\href{https://www.connections-answer.com/connections-difficulty-chart/}{Connections Difficulty Chart}}, providing ratings from 1 (easiest) to 5 (most challenging) by independent testers.

The resulting distribution of difficulty ratings for our dataset is provided in Appendix Figure \ref{fig:difficultyDistribution}. 

\subsection{Sample Heuristic} 
\begin{table}[ht]
    \centering
    \begin{tabular}{@{}lc@{}}
        \toprule
        Method & Performance\\
        \midrule
        $I$, $s$ and $V$ & 13.25\\
        $I$ only & 13.25\\
        $s$ and $V$ & 9.25\\
        \bottomrule
    \end{tabular}
    \caption{Ablation study for factors comprising our Group Similarity Score over 100 median difficulty \corpusname{} matches on the Multiple Tries configuration. $I$ is shown to be the most influential factor when choosing the best candidate solutions.}
    \label{fig:heuristicAblation}
\end{table}
To establish a baseline and demonstrate the challenge of the task, we designed a heuristic that mimics a player's initial, intuitive approach to the puzzles. It evaluates 4-word groups using a score $S = G - P$, where $G$ is group similarity and $P$ is a penalty for similarity to other words.

\paragraph{Group Similarity Score}
For a candidate group $C = \{c_1, c_2, c_3, c_4\}$, we compute $G$ as follows:

\begin{enumerate}[itemsep=-1pt, topsep=0pt, parsep=0pt, leftmargin=*]
    \item Obtain word embeddings $E$ using a pre-trained language model.
    \item Compute a clustering score $I = -K(E)$, where $K$ is the inertia (sum of squared distances to the centroid) of a k-means cluster (k=1).
    \item Calculate the minimum pairwise cosine similarity $s$ among words in the group.
    \item Compute a variance-based score $V = \frac{\text{mean}(P)}{1 + \text{var}(P)}$, where $P$ is the set of all pairwise similarities.
\end{enumerate}

The final score is a weighted sum: $G = 0.4 \cdot I + 0.3 \cdot s + 0.3 \cdot V$. These weights were chosen based on ablation studies as depicted in Table \ref{fig:heuristicAblation}, giving slight preference to the strongest contributing factor, $I$.

This formulation captures cluster tightness ($I$), minimum relatedness ($s$), and similarity consistency ($V$), mirroring intuitive judgments about word relationships typical of System 1 thinking.

\paragraph{Penalty Score}
To prevent overly generic groupings, we compute a penalty $P$ that measures how similar a candidate group is to remaining words:

\[
P = \frac{1}{|R|}\sum_{r \in R}\cos(\mu_C, r)
\]

where $\mu_C$ is the mean embedding of the candidate group $C$, and $R$ is the set of remaining words. A lower $P$ indicates a more distinct group.

\paragraph{Beam Search}
To balance between finding seemingly good initial groupings and maintaining some flexibility, we employ a beam search algorithm with a width of 10:

\begin{enumerate}[itemsep=-1pt, topsep=0pt, parsep=0pt, leftmargin=*]
    \item Initialize with an empty solution.
    \item For each step (up to 4 groups):
        \begin{enumerate}[itemsep=-1pt, topsep=0pt, parsep=0pt]
            \item Form all possible groups of 4 from remaining words.
            \item Compute $S = G - P$ for each new group.
            \item Retain the top 10 partial solutions based on cumulative score.
        \end{enumerate}
    \item Return the highest-scoring complete solution.
\end{enumerate}

This approach balances exploration of alternative groupings with a preference for high-scoring, seemingly obvious solutions. By design, it's prone to the misdirections built into the puzzles, serving as an effective baseline for comparison with more advanced reasoning methods.

\section{Experimental Setup}

\paragraph{Experiments} We analyzed 100 puzzles from our corpus, centered around the median difficulty rating of 3.0. This selection ensures consistent challenge across subjects and enables fair comparisons between LLMs and humans. Our full difficulty distribution is depicted in Appendix Figure \ref{fig:difficultyDistribution}.

\paragraph{Evaluation Settings} We tested under three conditions: (1) \textbf{One Try}: single attempt, scored 100 for success or 0 for failure; (2) \textbf{No Hints}: up to four re-tries; and (3) \textbf{Full Hints}: up to four re-tries with ``one away'' hints. For (2) and (3), scores represent the percentage of correct groups ($A=\{0,25,50,75,100\}$). Detailed examples are in Appendix Figures~\ref{fig:threeSetupsExplanation} and \ref{fig:fullHintsExample}.

\paragraph{Models} We evaluated six recent LLMs: Claude 3.5 Sonnet, GPT-4, GPT-4o, Gemini 1.5 Pro, LLaMA 3 70B Instruct, and LLaMA 3.1 405B Instruct \citep{anthropic2024claude, openai2023gpt4, openai2024gpt4o, geminiteam2024geminifamilyhighlycapable, touvron2023llama}. LLaMA models were used with the default temperature of 0.6; all others were used with a temperature of 0.5. Our heuristic used Multilingual-E5-Large-Instruct embeddings \citep{wang2024multilinguale5textembeddings}.

\paragraph{Prompts} Models received detailed background, instructions, and an example game, mirroring human participants' information. We used three prompting methods: Input-Output (IO), Chain-of-Thought (CoT) \citep{wei2022chain}, and CoT with Self-Consistency (CoT-SC) \citep{wang2023selfconsistency}. Detailed prompts are in Appendix Figures~\ref{fig:promptsIOOneTry},\ref{fig:promptsIOMultipleTries} \& \ref{fig:promptsCoTMultipleTries}.

\paragraph{Random Guess} We implemented a random guess baseline. The probability of correct random guessing is approximately $3.81 \times 10^{-7}$ (0.0000381\%)\footnote{Calculated as the inverse of $\frac{\binom{16}{4} \times \binom{12}{4} \times \binom{8}{4} \times \binom{4}{4}}{4!}$.}.

\paragraph{Human Evaluation} Three human evaluators completed 50 \textit{One Try}, 25 \textit{No Hints}, and 25 \textit{Full Hints} instances via a custom application. Performance was averaged across evaluators for each configuration.

\section{Results}

\paragraph{LLMs Significantly Underperform Humans}
As shown in Table \ref{tab:performance_accross_setups_small}, there is a substantial performance gap between LLMs and human participants across all testing configurations. Even in the most favorable scenario (Full Hints), the best-performing model, Claude 3.5, achieves only 40.25\% accuracy compared to humans' 60.67\%. This disparity becomes even more pronounced in more challenging setups. In the One Try configuration, the top LLM (Claude 3.5) manages only 11\% accuracy, while humans achieve 39.33\%. 

\begin{table}[t]
    \centering
    \small
    \begin{tabular}{@{}lccc@{}}
        \toprule
        Player & One Try & No Hints & Full Hints \\
        \midrule
        GPT-4  & 4.0 & 35.5 & 32.5 \\
        Claude 3.5 & \textbf{11.0} & 36.75 & \textbf{40.25} \\
        GPT-4o & 8.0 & \textbf{45.0} & 33.75 \\
        LLaMA 3.1 405b & 7.0 & 35.5 & 34.75 \\ 
        Gemini 1.5 Pro & 5.0 & 30.5 & 31.5 \\
        LLaMA 3 70b & 1.0 & 23.75 & 28.5 \\ 
        \midrule
        Random & 0.0 & 0.0 & 0.0\\
        Heuristic & 1.0 & 13.25 & 13.25 \\
        Humans & \phantom{*}\textbf{39.33}* & \phantom{*}\textbf{56.0}* & \phantom{*}\textbf{60.67}* \\
        \bottomrule
    \end{tabular}
    \vspace{-1mm}
    \caption{Performance (\%) of the models, baselines, and humans across our three setups, using IO prompting. * indicates statistical significance ($p<0.05$) of human performance compared to the top-performing model. See Appendix \ref{Stat-Sig} for statistical test  methodology.}
    \label{tab:performance_accross_setups_small}
\end{table}

\begin{figure}[ht]
    \centering
    \includegraphics[width=0.93\columnwidth]{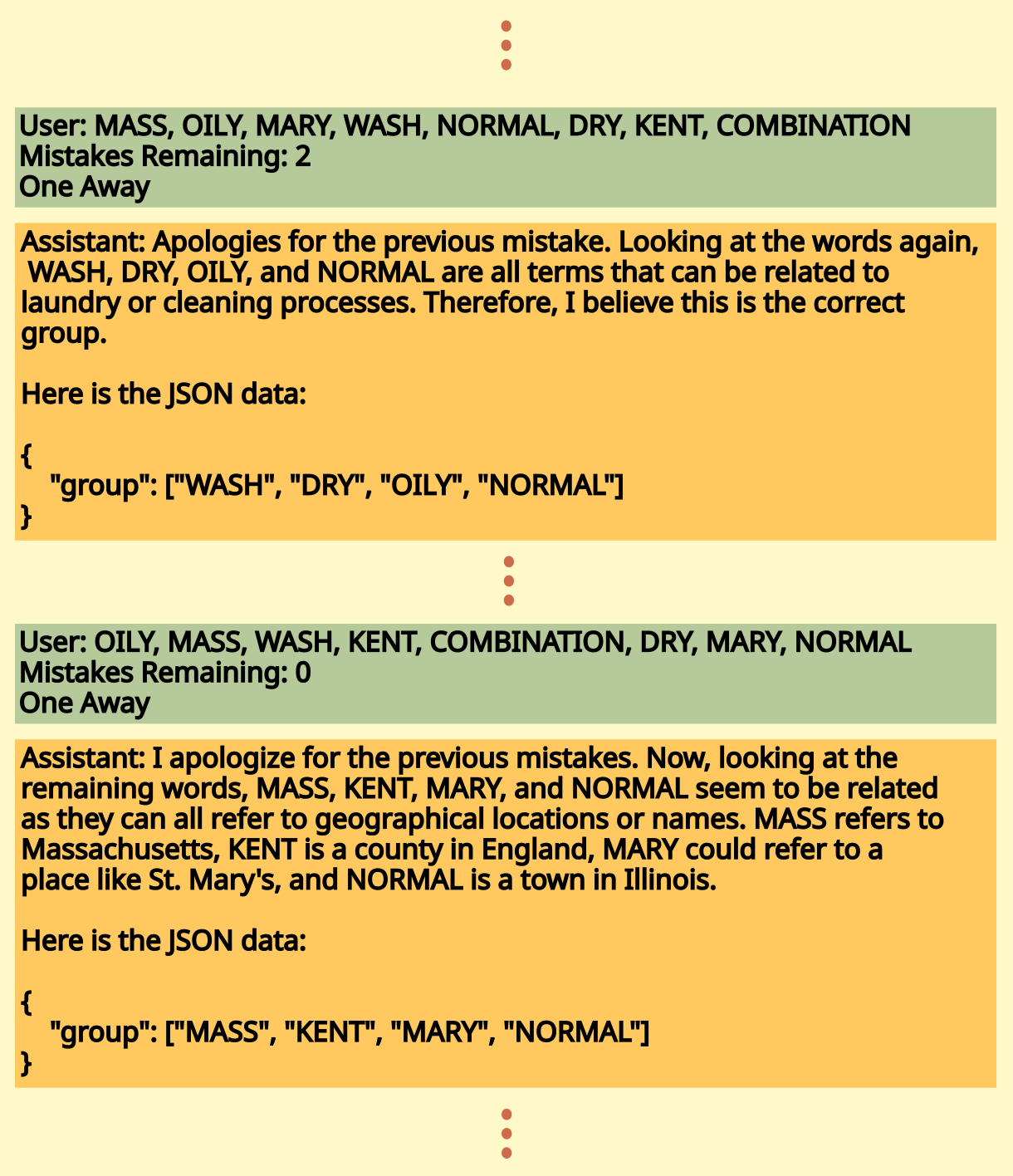}
    \caption{An example of GPT-4's output demonstrating the shallow reasoning of Chain-of-Thought-based approaches. The model first latches on to words in a laundry category, while in the second example, the model correctly identifies the group but fails to produce effective word groupings.}
    \label{fig:ShallowReasoning}
\end{figure}

\paragraph{Chain-of-Thought-Based Prompting Techniques Are Limited by Shallow Thinking} Figure~\ref{fig:ShallowReasoning} depicts two model outputs that illustrate the limited reasoning ability of Chain-of-Thought-based approaches. In this case, GPT-4 fails to consider multiple factors that may lead to better results --- such as when the remaining words outside of the chosen group contain strong matches --- the discovery of which requires more deliberate and specialized reasoning. This demonstrates the fundamental limitations of System 1 thinkers when performing non-symbolic reasoning tasks, even when endowed with complex methodology such as Chain-of-Thought, an issue that has been further explored in recent work \citep{sprague2024cotcotchainofthoughthelps}.

\begin{figure}[th]
	\centering
	\includegraphics[width=0.45\textwidth]
	{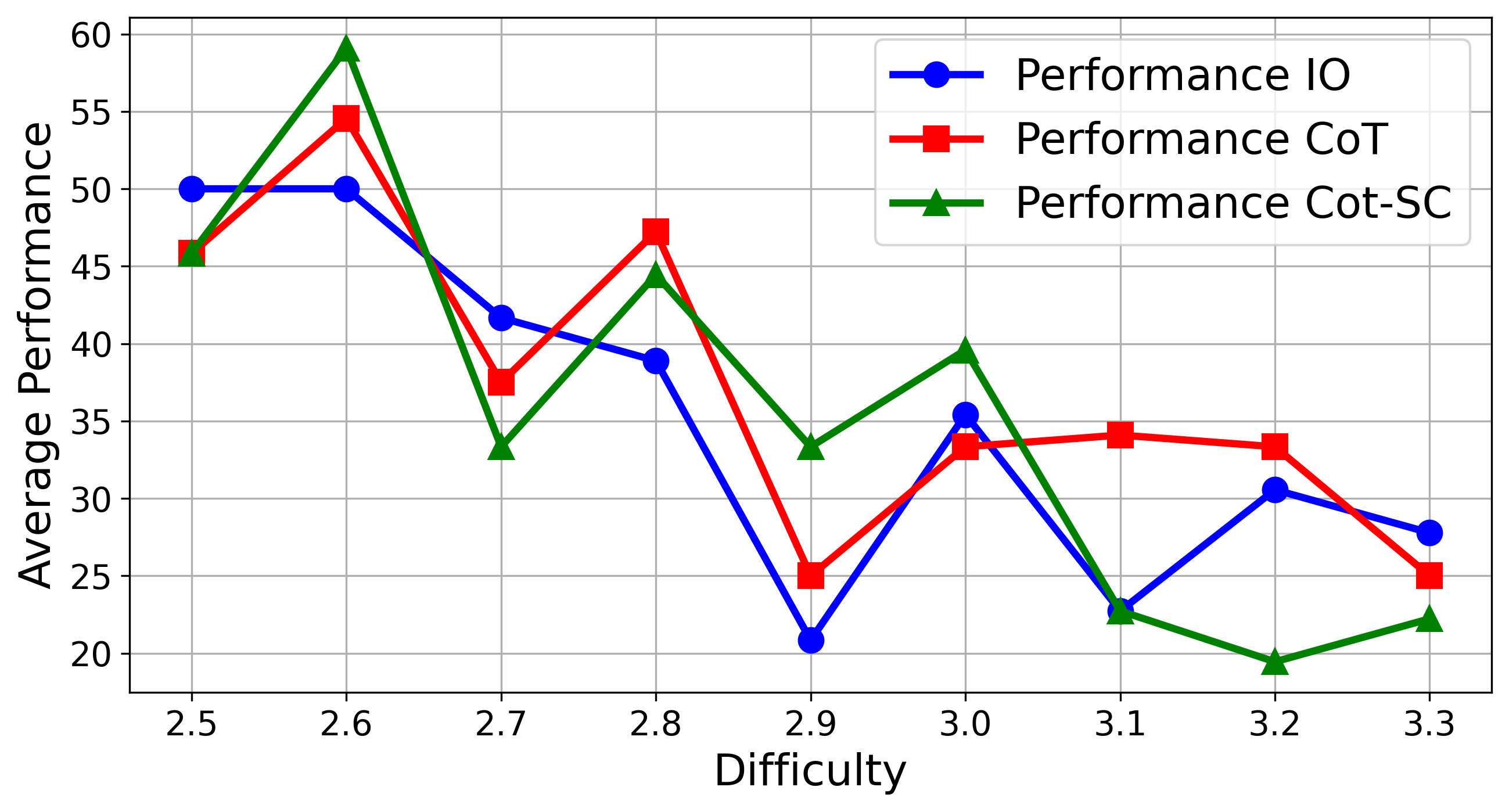}
 \vspace{-2mm}
	\caption{Average performance vs difficulty level for GPT-4 with various prompting techniques on \textit{Full Hints}}
	\label{fig:average_performance}
\end{figure}

\paragraph{Advanced Prompting Techniques Show Diminishing Returns}
Figure \ref{fig:average_performance} illustrates how  prompting methods such as Chain-of-Thought (CoT) and Self-Consistency (CoT-SC) do not consistently improve in performance as task difficulty increases. Surprisingly, simpler Input-Output (IO) prompting often outperforms these approaches, especially on harder puzzles. This suggests that current prompting techniques may be insufficient to simulate true System 2 reasoning in LLMs, and might even hinder performance by introducing unnecessary complexity.

\paragraph{Simple Heuristic is Comparable to Some LLMs}
As shown in Table \ref{tab:performance_accross_setups_small}, our baseline heuristic, designed to mimic intuitive System 1 thinking, achieves 13.25\% accuracy in both No Hints and Full Hints configurations. Notably, this performance is not far behind some of the tested LLMs, such as LLaMA 3 70b (23.75\% in No Hints, 28.5\% in Full Hints). This relatively small gap between a simple heuristic and more complex language models suggests that current LLMs demonstrate capabilities that fall between System 1-like pattern-matching and System 2-like deliberation, without fully achieving consistent, deliberate reasoning.

\paragraph{Contextual Hints of Limited Benefit to LLMs}
Referring again to Table \ref{tab:performance_accross_setups_small}, while human performance improves significantly with the addition of hints (from 56\% in No Hints to 60.67\% in Full Hints), few LLMs show meaningful improvements. Some models, like GPT-4o, paradoxically perform worse with full hints (33.75\%) compared to no hints (45\%). This suggests a fundamental difference in how LLMs and humans process and utilize contextual information in problem-solving tasks.

\paragraph{Performance Consistency Across Top LLMs}
Our evaluation reveals surprisingly consistent performance among top LLMs. Claude 3.5, GPT-4, and GPT-4o show no statistically significant differences from each other, a pattern that holds across all configurations. This suggests that \corpusname{} presents a unified challenge, exposing similar limitations in even the most advanced LLMs for tasks requiring System 2 thinking.

\section{Related Work}

Recent research has focused on developing benchmarks that address key challenges in evaluating language models' reasoning capabilities. Several works propose tasks that isolate specific cognitive processes, such as code reasoning, mathematical problem-solving, and logic \citep{liu2024codemind, mao2024champ, wu-etal-2024-reasoning}, aiming to disentangle task-specific knowledge from broader reasoning abilities. Researchers have also created benchmarks for deliberate, multi-step reasoning, including tasks designed to challenge System 1-style heuristics \citep{suzgun-etal-2023-challenging, mckenzie2024inversescalingbiggerisnt}. To combat data leakage, evolving datasets have been introduced, continuously updating with new problems from real-world sources \citep{sun-emami-2024-evograd, li2024evocodebench, jain2024livecodebench}. \corpusname{} uniquely combines these three aspects: isolating word relationship understanding, resisting simple heuristics, and maintaining novelty through regular updates.

The use of LLMs to solve \textit{Connections} has been contemporaneously explored by \citet{todd2024missedconnectionslateralthinking} and \citet{samdarshi-etal-2024-connecting}. \citet{todd2024missedconnectionslateralthinking} investigate the puzzle as a benchmark for abstract reasoning, examining sentence embedding baselines and analyzing solving strategies; \citet{samdarshi-etal-2024-connecting} analyze model performance across different knowledge categories, comparing these models to both novice and expert human performance benchmarks. We introduce three distinct evaluation configurations (One Try, No Hints, Full Hints), develop a concrete heuristic baseline that demonstrates the limitations of System 1 thinking, and maintain \corpusname{} as a living benchmark with regular updates to address data leakage concerns. This offers a novel framework specifically designed to assess LLMs' capacity for deliberate System 2 reasoning.

\section{Conclusion}
We introduced \corpusname, a benchmark that isolates word relationship understanding, penalizes heuristic-based thinking, and resists data leakage. Our evaluation of six LLMs, a simple heuristic, and human performance revealed significant gaps, with top models like GPT-4 falling nearly 30\% short of humans. This highlights the ongoing challenges in developing AI systems capable of deliberate reasoning. Future work should explore techniques to bridge this performance gap and investigate how improvements on \corpusname{} translate to other reasoning tasks.

\section*{Limitations}

\paragraph{Embedding Model Scale} Our heuristic uses a relatively small model due to hardware constraints. While this provides a baseline, it's possible that larger embedding models could yield different results. Future work should explore the scalability of our heuristic approach using more advanced embedding models to fully understand the relationship between model size and performance on \corpusname{}.

\paragraph{Prompt Engineering Scope} Cost constraints limited our ability to test an extensive range of prompting techniques. While we focused on standard, Chain-of-Thought, and Self-Consistency methods, future studies could explore a broader spectrum of prompting strategies, including more recent innovations. However, we intentionally excluded complex, long-context methods like Tree of Thoughts \cite{yao2023tree}, as these fall outside the scope of our focus on core reasoning capabilities.

\paragraph{Human Baseline Limitations} Our human performance data is derived from a small sample of three evaluators, which may not fully represent the broader population's problem-solving abilities. A larger-scale study with a diverse group of participants would provide a more robust human baseline and could reveal interesting patterns in human approaches to solving \textit{Connections} puzzles.

\paragraph{Temporal Limitations of the Dataset} While we commit to monthly updates of \corpusname{}, the dataset inherently represents a snapshot of puzzles from a specific time period. This could potentially limit its long-term relevance as language use and cultural references evolve. Regular assessments of the dataset's contemporary relevance may be necessary to maintain its effectiveness as a benchmark.

\paragraph{Cross-Cultural Applicability} The \textit{Connections} puzzles are primarily designed for an English-speaking, Western audience. This may limit the benchmark's applicability across different cultures and languages. Future work could explore creating multilingual versions or culturally adapted variants of \corpusname{} to assess LLM performance in more diverse contexts.
\section*{Acknowledgments}
This work was supported by the Natural Sciences and Engineering Research Council of Canada and by the New Frontiers in Research Fund. Angel Yahir Loredo Lopez was supported by the Mitacs Globalink Research Internship. Tyler McDonald was supported by the Natural Sciences and Engineering Research Council of Canada's Undergraduate Student Research Award.

\bibliography{custom}

\begin{thebibliography}{33}
\providecommand{\natexlab}[1]{#1}

\bibitem[{Anthropic(2024)}]{anthropic2024claude}
Anthropic. 2024.
\newblock Claude 3.5 sonnet model card addendum.
\newblock \url{https://www-cdn.anthropic.com/fed9cc193a14b84131812372d8d5857f8f304c52/Model_Card_Claude_3_Addendum.pdf}.

\bibitem[{Balloccu et~al.(2024)Balloccu, Schmidtov{\'a}, Lango, and Dusek}]{balloccu-etal-2024-leak}
Simone Balloccu, Patr{\'\i}cia Schmidtov{\'a}, Mateusz Lango, and Ondrej Dusek. 2024.
\newblock \href {https://aclanthology.org/2024.eacl-long.5} {Leak, cheat, repeat: Data contamination and evaluation malpractices in closed-source {LLM}s}.
\newblock In \emph{Proceedings of the 18th Conference of the European Chapter of the Association for Computational Linguistics (Volume 1: Long Papers)}, pages 67--93, St. Julian{'}s, Malta. Association for Computational Linguistics.

\bibitem[{Cobbe et~al.(2021)Cobbe, Kosaraju, Bavarian, Chen, Jun, Kaiser, Plappert, Tworek, Hilton, Nakano, Hesse, and Schulman}]{cobbe2021trainingverifierssolvemath}
Karl Cobbe, Vineet Kosaraju, Mohammad Bavarian, Mark Chen, Heewoo Jun, Lukasz Kaiser, Matthias Plappert, Jerry Tworek, Jacob Hilton, Reiichiro Nakano, Christopher Hesse, and John Schulman. 2021.
\newblock \href {https://arxiv.org/abs/2110.14168} {Training verifiers to solve math word problems}.
\newblock \emph{Preprint}, arXiv:2110.14168.

\bibitem[{Gautam et~al.(2024)Gautam, Bingert, Zhu, Lauscher, and Klakow}]{gautam2024robust}
Vagrant Gautam, Eileen Bingert, Dawei Zhu, Anne Lauscher, and Dietrich Klakow. 2024.
\newblock Robust pronoun use fidelity with english llms: Are they reasoning, repeating, or just biased?
\newblock \emph{arXiv preprint arXiv:2404.03134}.

\bibitem[{Geirhos et~al.(2020)Geirhos, Jacobsen, Michaelis, Zemel, Brendel, Bethge, and Wichmann}]{geirhos2020shortcut}
Robert Geirhos, J{\"o}rn-Henrik Jacobsen, Claudio Michaelis, Richard Zemel, Wieland Brendel, Matthias Bethge, and Felix~A Wichmann. 2020.
\newblock Shortcut learning in deep neural networks.
\newblock \emph{Nature Machine Intelligence}, 2(11):665--673.

\bibitem[{Gema et~al.(2024)Gema, Leang, Hong, Devoto, Mancino, Saxena, He, Zhao, Du, Madani et~al.}]{gema2024we}
Aryo~Pradipta Gema, Joshua Ong~Jun Leang, Giwon Hong, Alessio Devoto, Alberto Carlo~Maria Mancino, Rohit Saxena, Xuanli He, Yu~Zhao, Xiaotang Du, Mohammad Reza~Ghasemi Madani, et~al. 2024.
\newblock Are we done with mmlu?
\newblock \emph{arXiv preprint arXiv:2406.04127}.

\bibitem[{Hagendorff et~al.(2023)Hagendorff, Fabi, and Kosinski}]{Hagendorff_2023}
Thilo Hagendorff, Sarah Fabi, and Michal Kosinski. 2023.
\newblock \href {https://doi.org/10.1038/s43588-023-00527-x} {Human-like intuitive behavior and reasoning biases emerged in large language models but disappeared in chatgpt}.
\newblock \emph{Nature Computational Science}, 3(10):833–838.

\bibitem[{Hendrycks et~al.(2021{\natexlab{a}})Hendrycks, Burns, Basart, Zou, Mazeika, Song, and Steinhardt}]{hendrycks2021measuringmassivemultitasklanguage}
Dan Hendrycks, Collin Burns, Steven Basart, Andy Zou, Mantas Mazeika, Dawn Song, and Jacob Steinhardt. 2021{\natexlab{a}}.
\newblock \href {https://arxiv.org/abs/2009.03300} {Measuring massive multitask language understanding}.
\newblock \emph{Preprint}, arXiv:2009.03300.

\bibitem[{Hendrycks et~al.(2021{\natexlab{b}})Hendrycks, Burns, Kadavath, Arora, Basart, Tang, Song, and Steinhardt}]{hendrycks2021measuringmathematicalproblemsolving}
Dan Hendrycks, Collin Burns, Saurav Kadavath, Akul Arora, Steven Basart, Eric Tang, Dawn Song, and Jacob Steinhardt. 2021{\natexlab{b}}.
\newblock \href {https://arxiv.org/abs/2103.03874} {Measuring mathematical problem solving with the math dataset}.
\newblock \emph{Preprint}, arXiv:2103.03874.

\bibitem[{Huang et~al.(2024)Huang, Lin, Liu, Gong, Lu, Lei, Liang, Shen, Lin, Duan et~al.}]{huang2024competition}
Yiming Huang, Zhenghao Lin, Xiao Liu, Yeyun Gong, Shuai Lu, Fangyu Lei, Yaobo Liang, Yelong Shen, Chen Lin, Nan Duan, et~al. 2024.
\newblock Competition-level problems are effective llm evaluators.
\newblock In \emph{Findings of the Association for Computational Linguistics ACL 2024}, pages 13526--13544.

\bibitem[{Jain et~al.(2024)Jain, Han, Gu, Li, Yan, Zhang, Wang, Solar-Lezama, Sen, and Stoica}]{jain2024livecodebench}
Naman Jain, King Han, Alex Gu, Wen-Ding Li, Fanjia Yan, Tianjun Zhang, Sida Wang, Armando Solar-Lezama, Koushik Sen, and Ion Stoica. 2024.
\newblock Livecodebench: Holistic and contamination free evaluation of large language models for code.
\newblock \emph{arXiv preprint arXiv:2403.07974}.

\bibitem[{Li et~al.(2024)Li, Li, Zhang, Dong, and Jin}]{li2024evocodebench}
Jia Li, Ge~Li, Xuanming Zhang, Yihong Dong, and Zhi Jin. 2024.
\newblock Evocodebench: An evolving code generation benchmark aligned with real-world code repositories.
\newblock \emph{arXiv preprint arXiv:2404.00599}.

\bibitem[{Liu et~al.(2024)Liu, Zhang, and Jabbarvand}]{liu2024codemind}
Changshu Liu, Shizhuo~Dylan Zhang, and Reyhaneh Jabbarvand. 2024.
\newblock Codemind: A framework to challenge large language models for code reasoning.
\newblock \emph{arXiv preprint arXiv:2402.09664}.

\bibitem[{Liu(2023)}]{NYT2023connections}
Wyna Liu. 2023.
\newblock How our new game, connections, is put together.
\newblock \url{https://www.nytimes.com/2023/06/26/crosswords/new-game-connections.html}.
\newblock Accessed: 2024-09-09.

\bibitem[{Mao et~al.(2024)Mao, Kim, and Zhou}]{mao2024champ}
Yujun Mao, Yoon Kim, and Yilun Zhou. 2024.
\newblock Champ: A competition-level dataset for fine-grained analyses of llms' mathematical reasoning capabilities.
\newblock \emph{arXiv preprint arXiv:2401.06961}.

\bibitem[{McKenzie et~al.(2024)McKenzie, Lyzhov, Pieler, Parrish, Mueller, Prabhu, McLean, Kirtland, Ross, Liu, Gritsevskiy, Wurgaft, Kauffman, Recchia, Liu, Cavanagh, Weiss, Huang, Droid, Tseng, Korbak, Shen, Zhang, Zhou, Kim, Bowman, and Perez}]{mckenzie2024inversescalingbiggerisnt}
Ian~R. McKenzie, Alexander Lyzhov, Michael Pieler, Alicia Parrish, Aaron Mueller, Ameya Prabhu, Euan McLean, Aaron Kirtland, Alexis Ross, Alisa Liu, Andrew Gritsevskiy, Daniel Wurgaft, Derik Kauffman, Gabriel Recchia, Jiacheng Liu, Joe Cavanagh, Max Weiss, Sicong Huang, The~Floating Droid, Tom Tseng, Tomasz Korbak, Xudong Shen, Yuhui Zhang, Zhengping Zhou, Najoung Kim, Samuel~R. Bowman, and Ethan Perez. 2024.
\newblock \href {https://arxiv.org/abs/2306.09479} {Inverse scaling: When bigger isn't better}.
\newblock \emph{Preprint}, arXiv:2306.09479.

\bibitem[{OpenAI(2023)}]{openai2023gpt4}
OpenAI. 2023.
\newblock \href {https://arxiv.org/abs/2303.08774} {Gpt-4 technical report}.
\newblock \emph{Preprint}, arXiv:2303.08774.

\bibitem[{OpenAI(2024)}]{openai2024gpt4o}
OpenAI. 2024.
\newblock Hello gpt-4o.
\newblock \url{https://openai.com/index/hello-gpt-4o/}.
\newblock Accessed: 2024-09-04.

\bibitem[{Patel et~al.(2021)Patel, Bhattamishra, and Goyal}]{patel-etal-2021-nlp}
Arkil Patel, Satwik Bhattamishra, and Navin Goyal. 2021.
\newblock \href {https://doi.org/10.18653/v1/2021.naacl-main.168} {Are {NLP} models really able to solve simple math word problems?}
\newblock In \emph{Proceedings of the 2021 Conference of the North American Chapter of the Association for Computational Linguistics: Human Language Technologies}, pages 2080--2094, Online. Association for Computational Linguistics.

\bibitem[{Samdarshi et~al.(2024)Samdarshi, Mustafa, Kulkarni, Rothkopf, Chakrabarty, and Muresan}]{samdarshi-etal-2024-connecting}
Prisha Samdarshi, Mariam Mustafa, Anushka Kulkarni, Raven Rothkopf, Tuhin Chakrabarty, and Smaranda Muresan. 2024.
\newblock \href {https://doi.org/10.18653/v1/2024.emnlp-main.1182} {Connecting the dots: Evaluating abstract reasoning capabilities of {LLM}s using the {N}ew {Y}ork {T}imes connections word game}.
\newblock In \emph{Proceedings of the 2024 Conference on Empirical Methods in Natural Language Processing}, pages 21219--21236, Miami, Florida, USA. Association for Computational Linguistics.

\bibitem[{Sprague et~al.(2024)Sprague, Yin, Rodriguez, Jiang, Wadhwa, Singhal, Zhao, Ye, Mahowald, and Durrett}]{sprague2024cotcotchainofthoughthelps}
Zayne Sprague, Fangcong Yin, Juan~Diego Rodriguez, Dongwei Jiang, Manya Wadhwa, Prasann Singhal, Xinyu Zhao, Xi~Ye, Kyle Mahowald, and Greg Durrett. 2024.
\newblock \href {https://arxiv.org/abs/2409.12183} {To cot or not to cot? chain-of-thought helps mainly on math and symbolic reasoning}.
\newblock \emph{Preprint}, arXiv:2409.12183.

\bibitem[{Sun and Emami(2024)}]{sun-emami-2024-evograd}
Jing~Han Sun and Ali Emami. 2024.
\newblock \href {https://aclanthology.org/2024.lrec-main.592} {{E}vo{G}rad: A dynamic take on the {W}inograd schema challenge with human adversaries}.
\newblock In \emph{Proceedings of the 2024 Joint International Conference on Computational Linguistics, Language Resources and Evaluation (LREC-COLING 2024)}, pages 6701--6716, Torino, Italia. ELRA and ICCL.

\bibitem[{Suzgun et~al.(2023)Suzgun, Scales, Sch{\"a}rli, Gehrmann, Tay, Chung, Chowdhery, Le, Chi, Zhou, and Wei}]{suzgun-etal-2023-challenging}
Mirac Suzgun, Nathan Scales, Nathanael Sch{\"a}rli, Sebastian Gehrmann, Yi~Tay, Hyung~Won Chung, Aakanksha Chowdhery, Quoc Le, Ed~Chi, Denny Zhou, and Jason Wei. 2023.
\newblock \href {https://doi.org/10.18653/v1/2023.findings-acl.824} {Challenging {BIG}-bench tasks and whether chain-of-thought can solve them}.
\newblock In \emph{Findings of the Association for Computational Linguistics: ACL 2023}, pages 13003--13051, Toronto, Canada. Association for Computational Linguistics.

\bibitem[{Team et~al.(2023)Team, Anil, Borgeaud, Wu, Alayrac, Yu, Soricut, Schalkwyk, Dai, Hauth et~al.}]{geminiteam2024geminifamilyhighlycapable}
Gemini Team, Rohan Anil, Sebastian Borgeaud, Yonghui Wu, Jean-Baptiste Alayrac, Jiahui Yu, Radu Soricut, Johan Schalkwyk, Andrew~M Dai, Anja Hauth, et~al. 2023.
\newblock \href {https://arxiv.org/abs/2312.11805} {Gemini: A family of highly capable multimodal models}.
\newblock ArXiv preprint arXiv:2312.11805.

\bibitem[{{The New York Times}(2024)}]{NYTimes2024}
{The New York Times}. 2024.
\newblock Connections.
\newblock \url{https://www.nytimes.com/games/connections}.
\newblock Accessed: June 12, 2023 to June 03, 2024.

\bibitem[{Todd et~al.(2024)Todd, Merino, Earle, and Togelius}]{todd2024missedconnectionslateralthinking}
Graham Todd, Tim Merino, Sam Earle, and Julian Togelius. 2024.
\newblock \href {https://arxiv.org/abs/2404.11730} {Missed connections: Lateral thinking puzzles for large language models}.
\newblock \emph{Preprint}, arXiv:2404.11730.

\bibitem[{Touvron et~al.(2023)Touvron, Lavril, Izacard, Martinet, Lachaux, Lacroix, Rozi{\`e}re, Goyal, Hambro, Azhar et~al.}]{touvron2023llama}
Hugo Touvron, Thibaut Lavril, Gautier Izacard, Xavier Martinet, Marie-Anne Lachaux, Timoth{\'e}e Lacroix, Baptiste Rozi{\`e}re, Naman Goyal, Eric Hambro, Faisal Azhar, et~al. 2023.
\newblock Llama: Open and efficient foundation language models.
\newblock \emph{arXiv preprint arXiv:2302.13971}.

\bibitem[{Trichelair et~al.(2019)Trichelair, Emami, Trischler, Suleman, and Cheung}]{trichelair-etal-2019-reasonable}
Paul Trichelair, Ali Emami, Adam Trischler, Kaheer Suleman, and Jackie Chi~Kit Cheung. 2019.
\newblock \href {https://doi.org/10.18653/v1/D19-1335} {How reasonable are common-sense reasoning tasks: A case-study on the {W}inograd schema challenge and {SWAG}}.
\newblock In \emph{Proceedings of the 2019 Conference on Empirical Methods in Natural Language Processing and the 9th International Joint Conference on Natural Language Processing (EMNLP-IJCNLP)}, pages 3382--3387, Hong Kong, China. Association for Computational Linguistics.

\bibitem[{Wang et~al.(2024)Wang, Yang, Huang, Yang, Majumder, and Wei}]{wang2024multilinguale5textembeddings}
Liang Wang, Nan Yang, Xiaolong Huang, Linjun Yang, Rangan Majumder, and Furu Wei. 2024.
\newblock \href {https://arxiv.org/abs/2402.05672} {Multilingual e5 text embeddings: A technical report}.
\newblock \emph{Preprint}, arXiv:2402.05672.

\bibitem[{Wang et~al.(2023)Wang, Wei, Schuurmans, Le, Chi, Narang, Chowdhery, and Zhou}]{wang2023selfconsistency}
Xuezhi Wang, Jason Wei, Dale Schuurmans, Quoc~V Le, Ed~H. Chi, Sharan Narang, Aakanksha Chowdhery, and Denny Zhou. 2023.
\newblock \href {https://openreview.net/forum?id=1PL1NIMMrw} {Self-consistency improves chain of thought reasoning in language models}.
\newblock In \emph{The Eleventh International Conference on Learning Representations}.

\bibitem[{Wei et~al.(2022)Wei, Wang, Schuurmans, Bosma, Chi, Le, and Zhou}]{wei2022chain}
Jason Wei, Xuezhi Wang, Dale Schuurmans, Maarten Bosma, Ed~Chi, Quoc Le, and Denny Zhou. 2022.
\newblock Chain of thought prompting elicits reasoning in large language models.
\newblock \emph{arXiv preprint arXiv:2201.11903}.

\bibitem[{Wu et~al.(2024)Wu, Qiu, Ross, Aky{\"u}rek, Chen, Wang, Kim, Andreas, and Kim}]{wu-etal-2024-reasoning}
Zhaofeng Wu, Linlu Qiu, Alexis Ross, Ekin Aky{\"u}rek, Boyuan Chen, Bailin Wang, Najoung Kim, Jacob Andreas, and Yoon Kim. 2024.
\newblock \href {https://doi.org/10.18653/v1/2024.naacl-long.102} {Reasoning or reciting? exploring the capabilities and limitations of language models through counterfactual tasks}.
\newblock In \emph{Proceedings of the 2024 Conference of the North American Chapter of the Association for Computational Linguistics: Human Language Technologies (Volume 1: Long Papers)}, pages 1819--1862, Mexico City, Mexico. Association for Computational Linguistics.

\bibitem[{Yao et~al.(2023)Yao, Yu, Zhao, Shafran, Griffiths, Cao, and Narasimhan}]{yao2023tree}
Shunyu Yao, Dian Yu, Jeffrey Zhao, Izhak Shafran, Thomas~L Griffiths, Yuan Cao, and Karthik Narasimhan. 2023.
\newblock Tree of thoughts: Deliberate problem solving with large language models.
\newblock \emph{arXiv preprint arXiv:2305.10601}.

\end{thebibliography}

\clearpage
\onecolumn
\appendix
\section{Appendix}
\label{sec:appendix}

\subsection{Statistical Testing Procedure}
\label{Stat-Sig}
We used different statistical tests for the \textit{One Try} setup versus the \textit{No Hints} and \textit{Full Hints} setups due to the nature of the data in each case:

\textbf{One Try Setup:} We used a two-proportion $z$-test because the outcomes in this setup are binary (success or failure), making it appropriate for comparing two proportions.

\textbf{No Hints and Full Hints Setups:} For these setups, we used the Mann-Whitney U test because the outcomes are ordinal categorical data (number of correct groups, $A=\{0,1,2,3,4\}$). This non-parametric test is suitable for comparing the distribution of scores between two independent groups when the dependent variable is ordinal.

All tests were conducted at a 95\% confidence interval ($p<0.05$). We performed tests between the human evaluators and the top-performing model, as well as between the top two performing models, to assess the statistical significance of performance differences.

\subsection{Game Setup \& Prompting Details}
\label{sec:setup}
\begin{figure*}[ht]
    \centering
    \includegraphics[width=\textwidth]{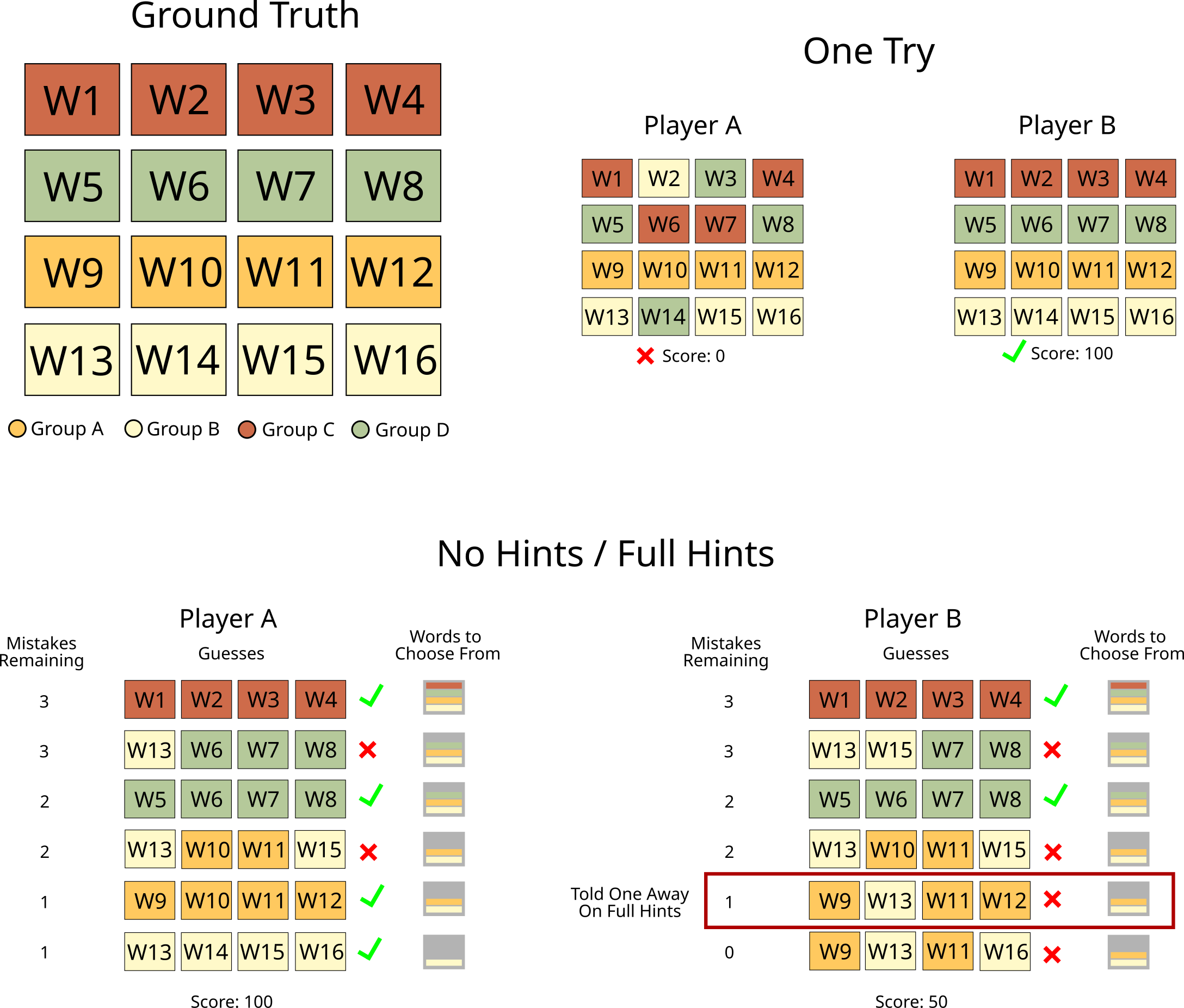}
    \caption{Demonstration of our three setups. \textbf{One Try}: Players have one chance to classify the words into the four groups. \textbf{No Hints}: Players have 4 chances to get the correct groups, where at each chance they are tasked to find a correct grouping. \textbf{Full Hints}: Same as No Hints, but players are told when they are one word away from a correct grouping.}
    \label{fig:threeSetupsExplanation}
\end{figure*}

\begin{figure*}[ht]
    \centering
    \includegraphics[width=\textwidth]{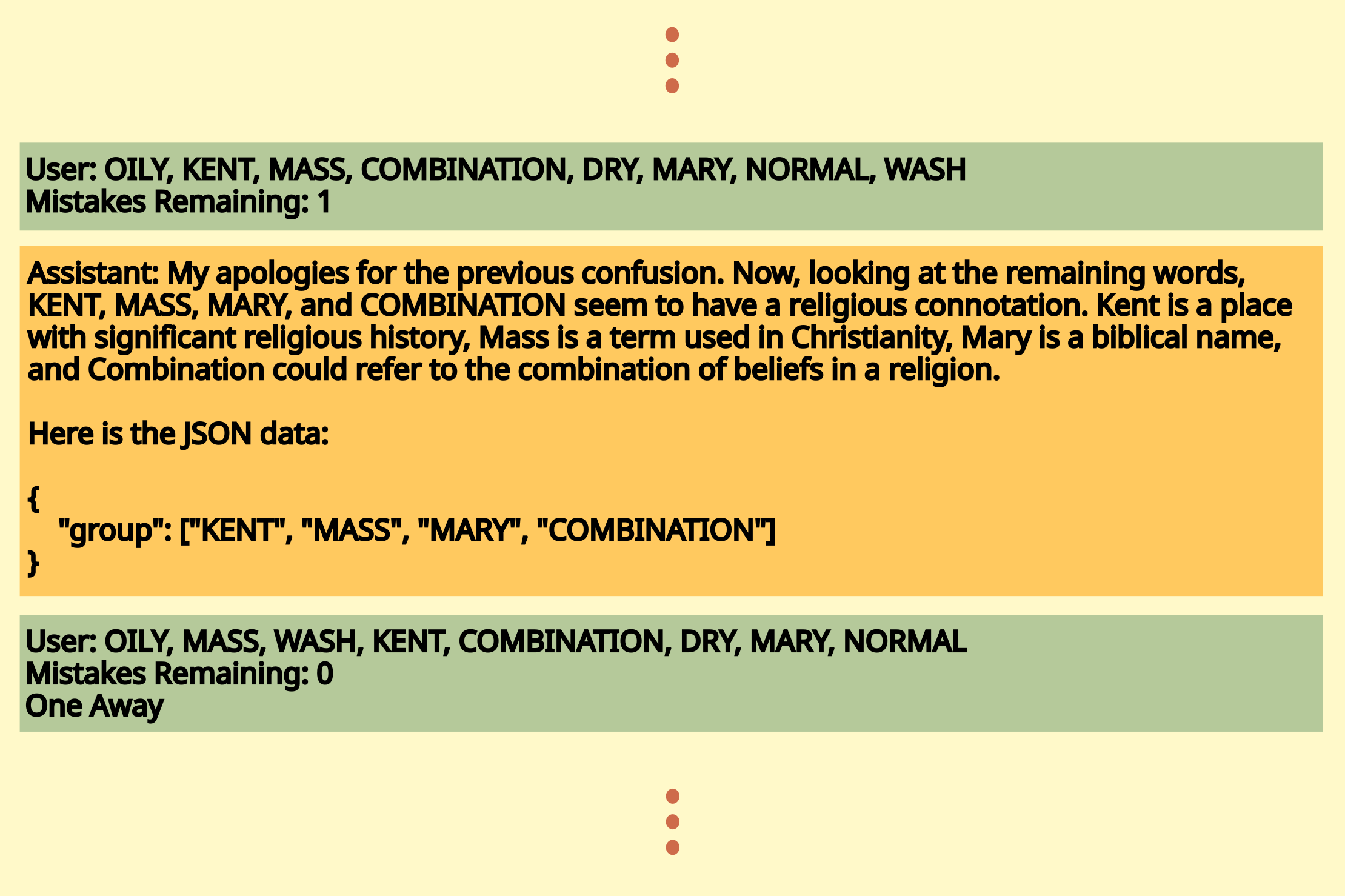}
    \caption{Example output from GPT-4 for Full Hints configuration, showing the ``One Away'' hint being given to the player.}
    \label{fig:fullHintsExample}
\end{figure*}

\begin{figure*}[ht]
    \centering
    \includegraphics[width=\textwidth]{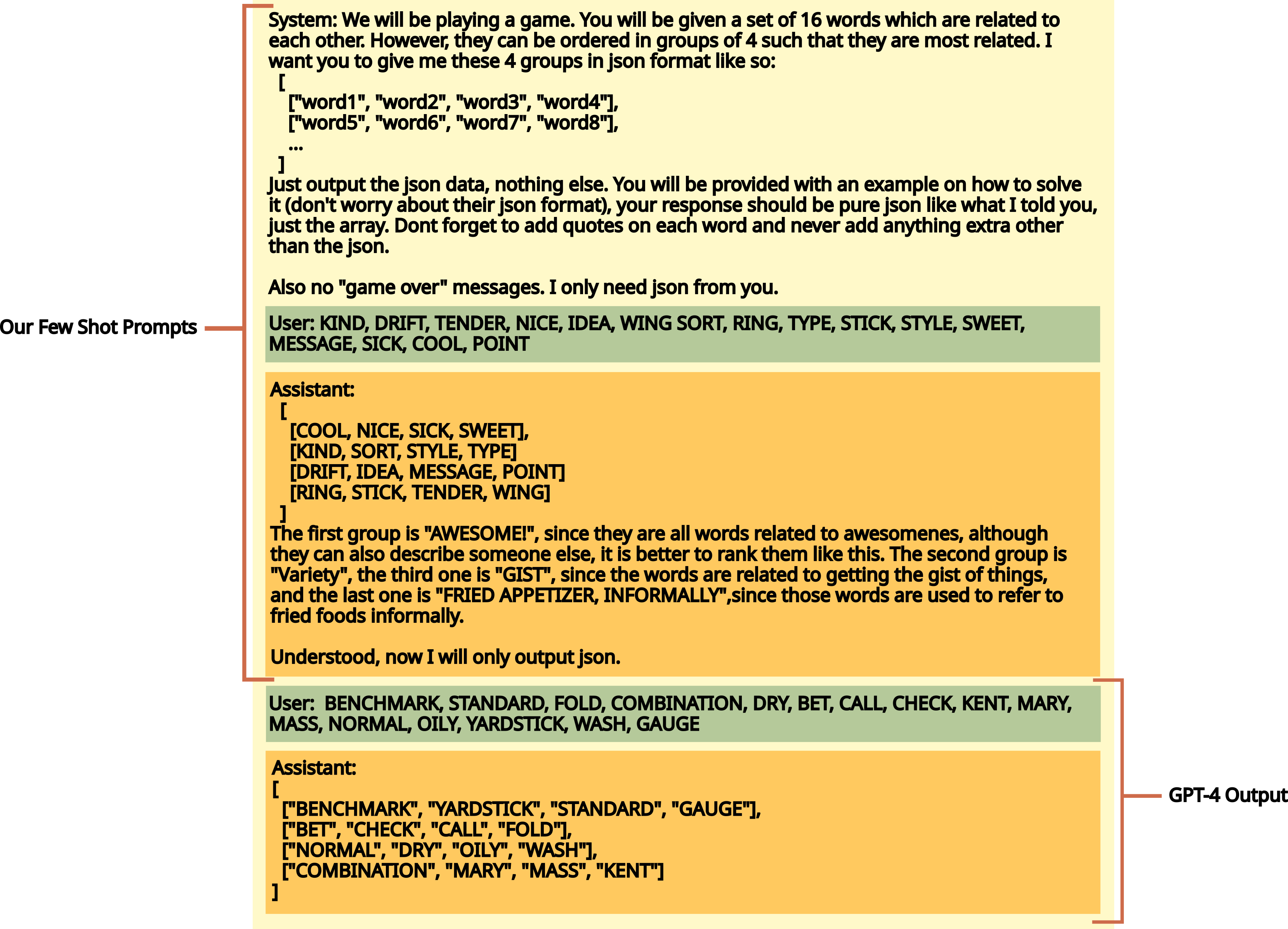}
    \caption{Prompts used for testing IO performance on the One Try setup, with GPT-4 example output.}
    \label{fig:promptsIOOneTry}
\end{figure*}

\begin{figure*}[ht]
    \centering
    \includegraphics[width=\textwidth]{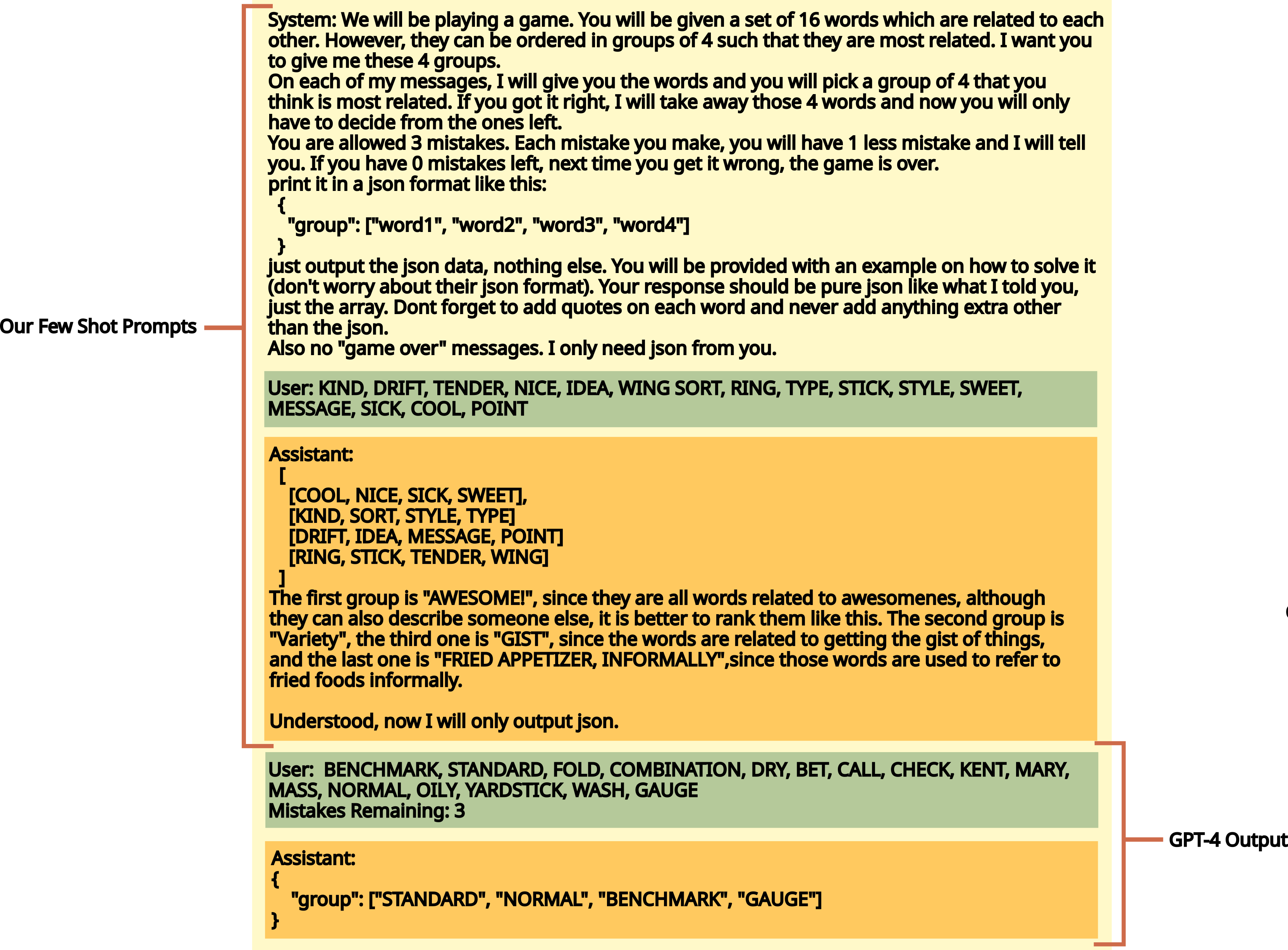}
    \caption{Prompts used for testing IO performance on the No Hints and Full Hints setups, with GPT-4 example output.}
    \label{fig:promptsIOMultipleTries}
\end{figure*}

\begin{figure*}[ht]
    \centering
    \includegraphics[width=\textwidth]{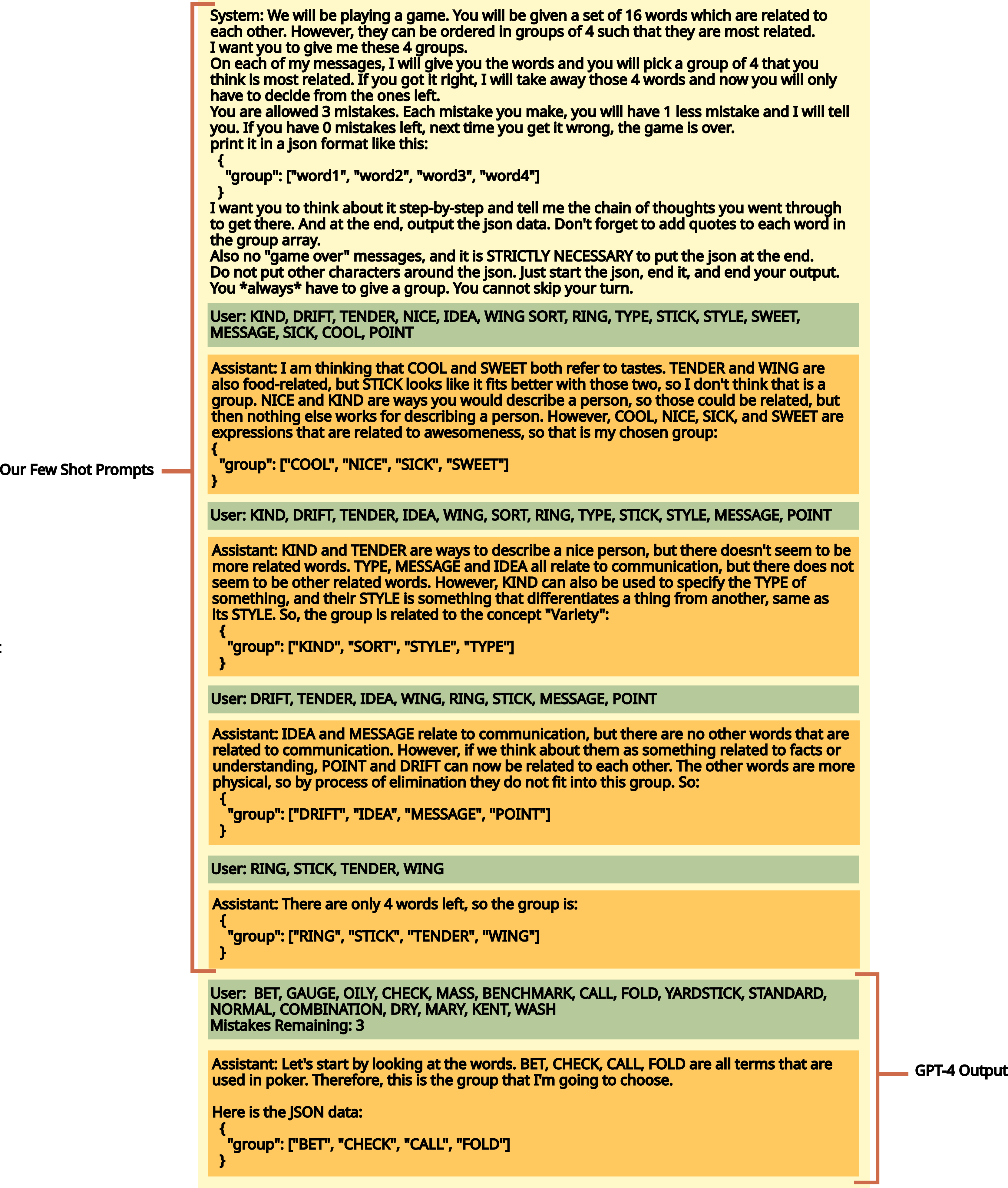}
    \caption{Prompts used for testing CoT performance on the No Hints and Full Hints setups, with GPT-4 example output.}
    \label{fig:promptsCoTMultipleTries}
\end{figure*}

\newpage
\clearpage

\subsection{Dataset Difficulty Composition}
\label{sec:sup}
\begin{figure*}[ht]
    \centering
    \includegraphics[width=\textwidth]{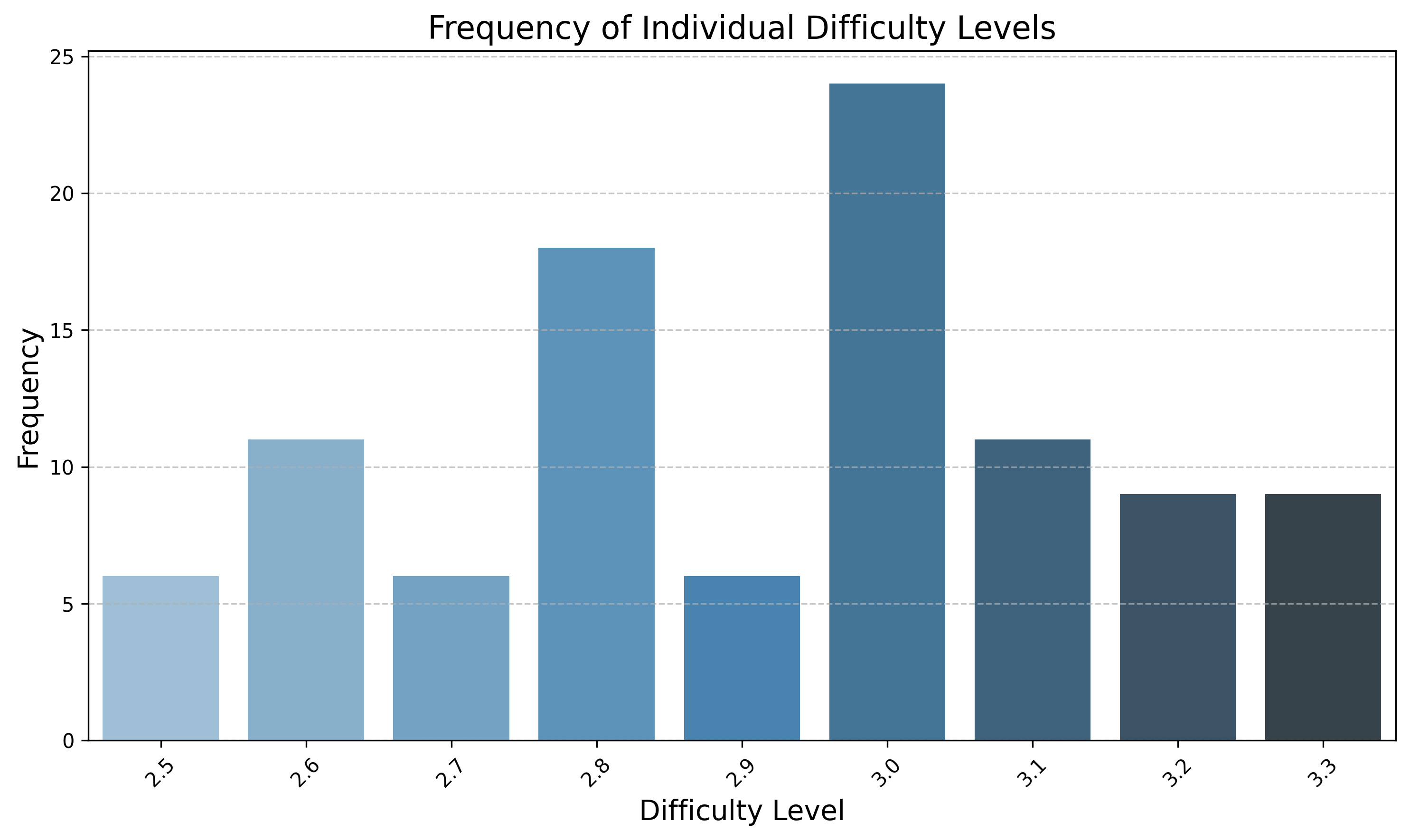}
    \caption{Difficulty distribution for our 100 median \corpusname{} instances.}
    \label{fig:difficultyDistribution}
\end{figure*}


\end{document}